\documentclass[10pt,twocolumn,letterpaper]{article}

\usepackage{wacv}
\usepackage{times}
\usepackage{epsfig}
\usepackage{graphicx}
\usepackage{amsmath}
\usepackage{amssymb}
\usepackage{balance}

\usepackage[pagebackref=true,breaklinks=true,letterpaper=true,colorlinks,bookmarks=false]{hyperref}

\wacvfinalcopy 

\def\wacvPaperID{415} 

\ifwacvfinal\fi
\begin{document}

\title{Automatic Tracker Selection w.r.t Object Detection Performance}


\author{Duc Phu Chau \and Fran\c cois Br\'emond \and Monique Thonnat \and Slawomir Bak \\
STARS team, INRIA, France \\
2004 route des Lucioles, 06560 Valbonne, France\\
{\tt\small \{Duc-Phu.Chau, Francois.Bremond, Monique.Thonnat, Slawomir.Bak\}  @inria.fr}
}

\maketitle

\begin{abstract}
   The tracking algorithm performance depends on video content. This paper presents a new multi-object tracking approach which is able to cope with video content variations. First the object detection is improved using Kanade-Lucas-Tomasi (KLT) feature tracking. Second, for each mobile object, an appropriate tracker is selected among a KLT-based tracker and a discriminative appearance-based tracker. This selection is supported by an online tracking evaluation. The approach has been experimented on three public video datasets. The experimental results show a better performance of the proposed approach compared to recent state of the art trackers.
\end{abstract}


\section{Introduction}

Many approaches have been proposed to track mobile objects in a scene. However the quality of tracking algorithms always depends on video content such as the crowded environment intensity or lighting condition. The selection of a tracking algorithm for an unknown scene becomes a hard task. Even when the tracker has already been determined, there are still some issues (\eg the determination of the best parameter values or the online estimation of the tracker quality) for adapting online this tracker to the video content variation.

Some approaches have been proposed to address these issues. For example, \cite{kuo} proposes an Adaboost-based algorithm for learning a discriminative appearance model for each mobile object. However the online Adaboost process is time consuming. Some other approaches propose to integrate different trackers and then select the appropriate tracker depending on video content. For example, \cite{prost} presents a framework which is able to select the most appropriate tracker among the three predefined trackers: normalized cross-correlation (NCC), mean-shift optical flow (FLOW) and online random forest (ORF). The approach is interesting but the online estimation of the tracker quality is not addressed. Also \cite{yoon12} proposes a tracking framework integrating multiple trackers based on different feature descriptors. All trackers are run in parallel. The output of each tracker is associated with a probability representing its quality. The framework selects the tracker corresponding to the highest probability for computing the tracking output. Both approaches require the execution of different trackers in parallel which is expensive in terms of processing time. Moreover 
the studies \cite{kuo, prost} take only into account the appearance variation of an object over time, but not tracking issues due to similar appearance of their neighboring objects.

In this paper, we propose a new object tracking approach overcoming the above limitations. The proposed strategy selects an appropriate tracker among an appearance tracker and a KLT tracker for each mobile object to obtain the best tracking performance. This helps to better adapt the tracking process to the spatial variation. Also, while the appearance-based tracker considers the object appearance, the KLT tracker takes into account the optical flow of pixels and their spatial neighbours. Therefore these two trackers can improve alternately the tracking performance. 

When spatially close objects have similar appearance, their tracking is more difficult. In order to solve this problem, for the appearance tracker, object descriptors are associated with discriminative weights while computing trajectories. These weights are updated automatically in function of the appearances of neighboring objects to ensure an enough discrimination between different tracked targets. This method does not require any training phase, neither parameter tuning, but still gets a robust object tracking performance. Object detection is also an issue when occlusions occur. Therefore, we also propose in this work a method to estimate object detection errors and correct them. This paper brings three following contributions: 

\begin{figure*}[]
\begin{center}
 \includegraphics[width=1\linewidth]{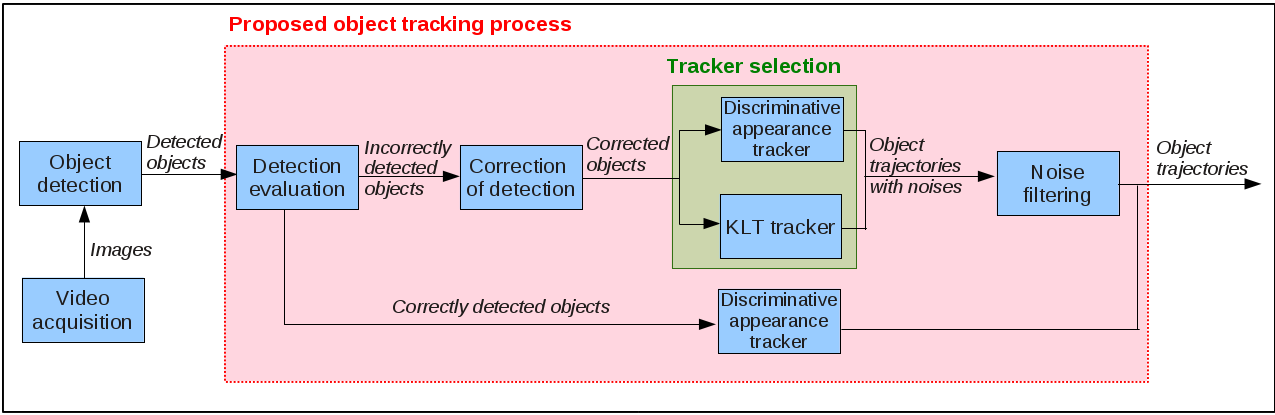}
\end{center}
   \caption{The scheme of the proposed approach}
\label{fig_approach}
\end{figure*}

\begin{itemize}
 \item An online evaluation for object tracking algorithms in videos
 \item An automatic tracker selection for optimizing the tracking performance
 \item A discriminative appearance tracker using object descriptor reliability
\end{itemize}

The paper is organized as follows. Section 2 details the proposed object tracking process. Section 3 is dedicated to the experimentation and validation of the proposed method. Section 4 presents concluding remarks.

\section{Proposed Object Tracking Process}

The proposed approach takes video images and object detection as input. For each frame, using a KLT feature tracker \cite{shi94}, we estimate whether an object is correctly detected. Correctly detected objects have in general reliable appearances. Therefore, we use an appearance tracker for tracking them over time. Incorrectly detected objects are regulated using the KLT feature tracking. For these objects, the tracking is performed by a tracker selected among the KLT and discriminative appearance trackers. This selection helps to ensure a reliable object matching. Figure \ref{fig_approach} presents the scheme of the proposed approach.

\subsection{Detection Evaluation}

When some mobile objects are too spatially close or occluded, the detection can fail because their appearances could be partially visible. In this work, we address this problem using a KLT feature tracking. For each object detected at $t$, if it overlaps more than one object detected at $t-1$, we label the KLT feature points belonging to this object. The KLT features coming from a same object at $t-1$ are labeled the same value. Objects that have more than one label is considered as ``incorrectly detected''. These objects can contain more than one object inside their bounding boxes. They are then corrected by the correction of detection task. The other objects are considered as correctly detected.

\subsection{Correction of Detection}
\label{sec_correction_detection}

The correction of detection is performed based on the label values of the KLT feature points. For KLT feature points having the same label value, a bounding box is created to cover them. So an original object is then split into smaller bounding boxes. In general, these bounding boxes are smaller than the real objects because some KLT points are not tracked. Therefore, the bounding box sizes are regulated according to the sizes of corresponding objects at instant $t-1$.

\begin{figure}[t]
\begin{center}
   \includegraphics[trim=0 30 0 35, clip, width=0.7\linewidth]{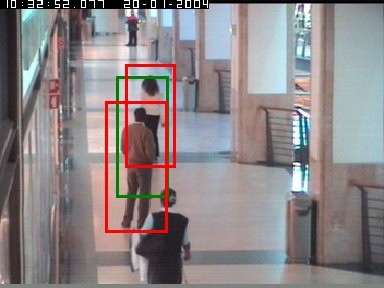}
\end{center}
   \caption{Illustration of the object detection correction for Caviar video. The green bounding box is the output of the object detection process. The red bounding boxes are the result of the detection correction task. }
\label{fig_split}
\end{figure}

Figure \ref{fig_split} illustrates the output of the correction of detection. The green bounding box is the output of the object detection task which covers two mobile objects. Using the KLT feature tracker, the correction of detection task splits the green bounding box into two bounding boxes (the red ones) and re-sizes them correctly.

\subsection{Discriminative Appearance Tracker}

In this paper, we propose an appearance tracker which relies on the coherence of five object appearance descriptors: 2D shape ratio, 2D area, color histogram,  color covariance and dominant color. Each object descriptor is effective for different cases. The descriptors concerning size as shape ratio, area can be used when only mobile object sizes are different each other. When the sizes of objects are similar, the color descriptors can be helpful to discriminate tracked objects. When the lighting condition of scene is not good, the color covariance descriptor can give a better object discrimination than color histogram and dominant color descriptors.

This tracker is composed of two stages. First, given an object $i$ detected at $t$, denoted $o_t^i$, and an object $j$ detected at $t-n$, denoted $o_{t-n}^j$, we define a similarity score for each descriptor $k$ ($k=1..5$) (section \ref{sec_desc_simil}). Second, the global similarity score between $o_t^i$ and $o_{t-n}^j$ is defined as a weighted combination of the five descriptor similarities (section \ref{sec_global_simil}). Successive links form several paths on which an object can undergo within the considered temporal window. Each possible path of an object is associated with a score given by all global similarities associated with the links it contains. The object trajectory is determined using the Hungarian algorithm.

\subsubsection{Descriptor Similarity}
\label{sec_desc_simil}

At each frame, for each object, the five object appearance descriptors are computed as follows.

 \textbf{- 2D Shape ratio:} Ratio between the width and height of the 2D bounding box of the object.
 
 \textbf{- 2D Area:} Area of the 2D bounding box of the object.

 \textbf{- Color histogram:} A normalized RGB color histogram of moving pixels inside the object bounding box.
 
 \textbf{- Color covariance:} In this paper, we use the covariance descriptor proposed in \cite{sbak}. For each pixel belonging to the object, we compute the following features: locations, RGB channel values, gradient magnitude and orientation in each channel. All computed feature values are then combined to define the color covariance descriptor of the object.
 
 \textbf{- Dominant color:} The dominant color descriptor is defined in \cite{yangCol}. This descriptor is similar to the color histogram descriptor, but it takes into account only the important colors of the object.
 
For each descriptor, we define a similarity score between $o_t^i$ and $o_{t-n}^j$, denoted $DS_k (o_t^i,\ o_{t-n}^j )$. For the 2D shape ratio and 2D area descriptors, the descriptor similarity is defined as the ratio between considered object descriptors. For the color histogram and dominant color descriptors, the earth mover distance (EMD) is used to compare two object histograms. The color covariance similarity is defined in function of distance between two covariance matrices proposed in \cite{forstner99}. The similarity corresponding to the color histogram, dominant color and color covariance descriptors is combined with a pyramid matching kernel to handle object occlusions.

\subsubsection{Global Similarity with Discriminative Descriptors}
\label{sec_global_simil}

The global similarity score between  $o_t^i$ and $o_{t-n}^j$ is defined as a weighted combination of the five descriptor similarities. However it is difficult to estimate the object descriptor weights because they depend on several elements such as lighting condition, density of mobile objects, object appearance. In this paper, we propose to use a discriminative technique to compute these weights. The descriptor weights have to be able to discriminate correctly the appearance between spatially close objects. This helps to reduce the object identity switch which is a common drawback in the tracking task. Therefore, in our approach, the descriptor weights are set differently for different objects, depending on their locations and appearances. Given an object $o_t^i$, the weight of descriptor $k$ for this object is defined as follows:
\begin{equation}
 w_k^i = \frac{1}{|\mathfrak{N}(o_t^i)|} \sum_{j=1}^{|\mathfrak{N}(o_t^i)|} lg \frac{1}{DS_k(o_t^i,\ o_t^j)},\ \ o_t^j \in  \mathfrak{N}(o_t^i)
\end{equation}

\noindent where $ \mathfrak{N}(o_t^i)$ is a set of neighboring objects of $o_t^i$ at time $t$ and is defined as $\mathfrak{N}(o_t^i) = \{o_t^j/\ j \neq i \ \wedge \ dist\_2D(o_t^i, o_t^j) < \epsilon_1 \ \wedge \ dist\_3D(o_t^i, o_t^j) < \epsilon_2 \} $; $dist\_2D(o_t^i, o_t^j)$ and $dist\_3D(o_t^i, o_t^j)$ be the 2D and 3D distances between $o_t^i$ and $o_t^j$; $\epsilon_1$ and $\epsilon_2$ are two predefined thresholds. Logarithm is an increasing function. Lower the descriptor similarity score between the spatially close objects is, the higher the object descriptor weight. 

Using the descriptor weights determined by this discriminative method, the global similarity score between $o_t^i$ and $o_{t-n}^j$, denoted $GS(o_t^i, o_{t-n}^j)$ is defined:
\begin{equation}
\label{eq_simil}
GS(o_t^i, o_{t-n}^j) = \frac{\sum_{k=1}^{5} (w_k ^i + w_k ^j ) DS_k(o_t^i, o_{t-n}^j)}{\sum_{k=1}^{5} (w_k ^i + w_k ^j )}
\end{equation}

A tracked object is defined as ``inactivated'' if it is not matched with any object detected at the current frame $t$. We construct a matrix $\mathbf{M} = \{m_{ij} \}$, with $i = 1.. r,\ j = 1..c $ where $r$ represents the number of detected objects at $t$ and $c$ represents the number of inactivated tracked objects in a given temporal window $[t-T,\ t - 1]$; $m_{ij}$ represents the global similarity score between object $o_t^i$ and $o_{t-n}^j$. The object tracking problem is now transformed to the assignment problem which has to optimize the sum of matching scores. In this paper, the Hungarian algorithm is used to solve this problem.

\subsection{KLT Tracker}
\label{sec_klt}

The second mobile object tracker relies on the tracking of Kanade-Lucas-Tomasi (KLT) features \cite{shi94}. The KLT tracker takes the detected objects as input. This tracker includes three steps: KLT feature selection, KLT feature tracking and object tracking. 

The objective of the KLT feature selection is to detect the KLT features located on the detected objects using the eigenvalues of their gradient matrices. In the KLT feature tracking step, each KLT feature is tracked by optimizing the translation of its feature point window. 

The objective of the object tracking step is to compute object trajectories. This task relies on the number of matching feature points over frames between detected objects. Let $P$ be the number of matching KLT features between two objects $o_t^i$ and $o_{t-1}^j$. We define a similarity score between these two objects as follows:
\begin{equation}
S_{KLT} (o_t^i, o_{t-1}^j) = min (  \frac{P}{M_{o_{t-1}^j}},  \frac{P} {M_{o_t^i}} )
\end{equation}

\noindent where $M_{o_{t-1}^j}$ and $M_{o_t^i}$ are respectively the total number of KLT feature points located on object $o_{t-1}$ and $o_t$. The Hungarian algorithm is then applied to find the best matching of objects between $t-1$ and $t$.


\subsection{Tracker Selection}


For objects which have a spatial overlap, it is difficult to decide which tracker can be more appropriate to track them. The discriminative appearance tracker can fail because the appearance is not fully visible. The KLT tracker can fail if the number of matching KLT features is not efficient or the KLT features are located on an image background. Therefore we propose a tracker selection based on an online tracking evaluation.

\subsubsection{Online Tracking Evaluation}

The state of a mobile object $o_t^i$ at instant $t$ is defined as $\{ x, y, \mathcal{W}, \mathcal{H}  \}  $ where the first two elements represent the object 2D coordinates, the last two elements represent width and height of its 2D bounding box. The observation of an object is defined as a set of five appearance descriptors: 2D shape ratio, 2D area, color histogram, color covariance and dominant color. Given an ``inactivated'' tracked object $o_{t-n}^j$, matching an object at frame $t$, denoted $\widehat{ o_t^j}$, is supposed to maximize the joint probability distribution:
\begin{equation}
\label{eq_best_obj}
\widehat{ o_t^j} = \underset{o_t^i}{ \operatorname{argmax}} \ \mathcal{P} (o_t^i, \{ \mathcal{M}_k^{o_{t-n}^j } \}_{k=1..5} )
\end{equation}

\noindent where $\mathcal{M}_k^{o_{t-n}^j }$ is the model of the appearance descriptor $k$ for object $o_{t-n}$. These descriptor models represent the observation of $o_{t-n}^j$ of its last $Q$ frames ($Q$ is a predefined parameter). Assuming the independence of these five appearance models, we obtain:
\begin{equation}
\label{eq_pro_descs}
 \mathcal{P} (o_t^i, \{ \mathcal{M}_k^{o_{t-n}^j } \}_{k=1..5} ) = \overset{5} { \underset{k = 1} {\Pi} } \mathcal{P} (o_t^i | \mathcal{M}_k^{o_{t-n}^j} )
\end{equation}

\noindent where $\mathcal{P} (o_t^i | \mathcal{M}_k^{o_{t-n}^j} )$ represents the probability of object (or candidate region) $o_t^i$ belonging to model $\mathcal{M}_k^{o_{t-n}^j}$ (of an already tracked object). This measures a general tracking reliability and can be computed for any tracker.

In the following section, we present how to compute $\mathcal{P} (o_t^i | \mathcal{M}_k^{o_{t-n}^j} )$ for the five descriptors.
\newline

\textbf{- 2D Shape Ratio and 2D Area (k = 1 and k = 2) }

By assuming that the variation of the 2D area and shape ratio of a tracked object follows a Gaussian distribution, we can use the Gaussian probability density function (PDF) to compute this score. Also, longer the trajectory of $o_{t-n}$ is, more reliable the PDF is. Let $\mathfrak{T}_{t-n}^j$ be the trajectory of $o_{t-n}^j$. For these two descriptors, we define the probability of an object $o_t^i$ belonging to the model $\mathcal{M}_k^{o_{t-n}^j}$ ($k = 1$ for 2D shape ratio and $k=2$ for 2D area descriptor) as follows:
\begin{equation}
\label{eq_pro_2d}
\mathcal{P} (o_t^i | \mathcal{M}_k^{o_{t-n}^j} ) = \frac{exp({-\frac{(s^k_{o_t^i} - \mu^k_{ \mathfrak{T}_{t-n}^j  } )^2  }  {2(\sigma^k_{\mathfrak{T}_{t-n}^j})^2}} ) }{\sqrt{2\pi\sigma_{\mathfrak{T}_{t-n}^j}^2}} min (\frac { | \mathfrak{T}_{t-n}^j | }{Q}, 1)
\end{equation}

\noindent where $s^k_{o_t^i}$ is the value of descriptor $k$ for object $o_t^i$ ($s^k_{o_t^i}$ can be 2D area or 2D shape ratio value), $\mu^k_{\mathfrak{T}_{t-n}^j}$ and $\sigma^k_{\mathfrak{T}_{t-n}^j}$ are respectively mean and standard deviation values of descriptor $k$ of last $Q$-objects belonging to $\mathfrak{T}_{t-n}^j$; $ | \mathfrak{T}_{t-n}^j |$ is time length (in number of frames) of $\mathfrak{T}_{t-n}^j$. By selecting the last $Q$-objects, the probability takes into account the latest variations of the $\mathfrak{T}_{t-n}^j$.
\newline

\textbf{- Color Histogram (k = 3) }

For each color channel $\alpha$ (\ie Red, Green or Blue), we compute the mean histogram representing the intensity of the last $Q$ detected objects belonging to $\mathfrak{T}_{t-n}^j$, denoted $\overline{H}^\alpha_{\mathfrak{T}_{t-n}^j}$. The probability of an object $o_t^i$ belonging to the model $\mathcal{M}_3^{o_{t-n}^j}$ is defined in function of similarities  between mean histograms and color histogram of $o_t^i$:

\begin{equation}
\label{eq_pro_hist}
\mathcal{P} (o_t^i | \mathcal{M}_3^{o_{t-n}^j} ) = \frac { \underset{\alpha  = R, G, B} {\sum} E (H^{\alpha}_{o_t^i},\ \overline{H}^\alpha_{\mathfrak{T}_{t-n}^j}) } {3} min (\frac{ | \mathfrak{T}_{t-n}^j | }{Q}, 1)
\end{equation}

\noindent where $H^{\alpha}_{o_t^i}$ represents the color histogram of $o_t^i$, $E$ is the earth mover distance. Similar to the formula \ref{eq_pro_2d}, longer the $\mathfrak{T}_{t-n}^j$ is, more reliable this probability is. Therefore, in the formula \ref{eq_pro_hist}, we also multiply the expression with $min$ $(\frac{ | \mathfrak{T}_{t-n}^j | } {Q}, 1)$.
\newline

\textbf{- Color Covariance (k = 4) }

Similar to the color histogram probability, for a trajectory $\mathfrak{T}_{t-n}^j$  we compute a mean color covariance matrix of $Q$ last objects belonging to $\mathfrak{T}_{t-n}^j$. The model probability of color covariance descriptor $\mathcal{P} (o_t^i | \mathcal{M}_4^{o_{t-n}^j} )$ is then defined in function of the similarity between the covariance matrix of $o_t^i$ and the mean covariance matrix of $\mathfrak{T}_{t-n}^j$.
\newline

\textbf{- Dominant Color (k = 5) }

First, we compute the descriptor similarity of dominant color $DS_5(o_t^i, o^q)$ between $o_t$ and each object $o^q$ belonging to last $Q$ objects of $\mathfrak{T}_{t-n}^j$.  The probability of an object $o_t^i$ belonging to the model $\mathcal{M}_5^{o_{t-n}^j}$ is defined as follows:
\begin{equation}
 \mathcal{P} (o_t^i | \mathcal{M}_5^{o_{t-n}^j} ) =  \frac{ \sum_{q=1}^Q  DS_5(o_t, o^q)}{Q}  min (\frac{T({\mathfrak{T}_{t-n}^j} )}{Q}, 1)
\end{equation}

\subsubsection{Tracker Selection}

In this paper, we propose to select the appropriate tracker among the discriminative appearance tracker (denoted $\mathbf{T}^1$) and KLT tracker (denoted $\mathbf{T}^2$). At instant $t$, for an inactivated tracked object $o_{t-n}^j$, the selected tracker $\mathbf{T}^{\hat{\beta } }$ is determined as follows:
\begin{equation}
 \hat{\beta} = \underset{\beta}{\operatorname{argmax}} \  \mathcal{P} (o_t^i, \{ \mathcal{M}_k^{o_{t-n}^j } \}_{k=1..5}  , \mathbf{T} = \mathbf{T}^\beta)
\end{equation}

\noindent where $\mathcal{P} (o_t^i, \{ \mathcal{M}_k^{o_{t-n}^j } \}_{k=1..5}  , \mathbf{T} = \mathbf{T}^\beta) $ represents the probability $\mathcal{P} (o_t^i, \{ \mathcal{M}_k^{o_{t-n}^j } \}_{k=1..5}$) while using $\mathbf{T}^\beta$. Given an inactivated tracked object $o_{t-n}^j$, a tracker is selected if this tracker proposes to link $o_{t-n}^j$ to an object $o_t^i$ maximizing the equation (\ref{eq_pro_descs}). When both trackers loose an object, the approach assumes that an occlusion or miss detection has occurred. In this case, the tracking is suspended and tracker waits for new detections. If new detections are matched with the suspended objects, tracking is resumed.

\subsection{Noise Filtering}

Among objects created by the split process (see section \ref{sec_correction_detection}), if an object is not selected according to equation (\ref{eq_best_obj}), it is considered as noise. This noise category appears when the KLT features linking to this object are not good (\eg KLT features located on image background, wrong feature linking). The noisy objects are removed from the tracking output.

\section{Experimental Results}

We experiment the proposed object tracking approach on three public video datasets: PETS 2009\footnote{http://www.cvg.rdg.ac.uk/PETS2009/a.html}, Caviar\footnote{http://homepages.inf.ed.ac.uk/rbf/CAVIARDATA1/} and TUD\footnote{http://www.d2.mpi-inf.mpg.de/node/428/}. A HOG-based algorithm combining with background subtraction \cite{corvee10} is used for detecting people. For each dataset, we present the tracking results of the appearance tracker (the object appearance descriptors  have the same weights), the KLT tracker and the proposed approach (combine both KLT tracker and the discriminative appearance tracker).

\subsection{PETS Dataset}

In this test, we use the tracking evaluation metrics presented in \cite{kasturi09} to compare with other tracking algorithms. The first metric is MOTA computing multiple object tracking accuracy. The second metric is MOTP computing multiple object tracking precision.  All these metrics are normalized in the interval $[0,\ 1]$. The higher these metrics, the better the tracking quality is.

The video of this test belongs to the PETS dataset 2009. We select the sequence S2\_L1, camera view 1, time 12.34 for testing because this sequence is experimented in several state of the art trackers. This sequence has 794 frames, contains 21 mobile objects and several occlusion cases.

\begin{figure*}[]
\begin{center}$
\begin{array}{ccc}
\includegraphics[trim=0 100 0 55, clip, width=5cm]{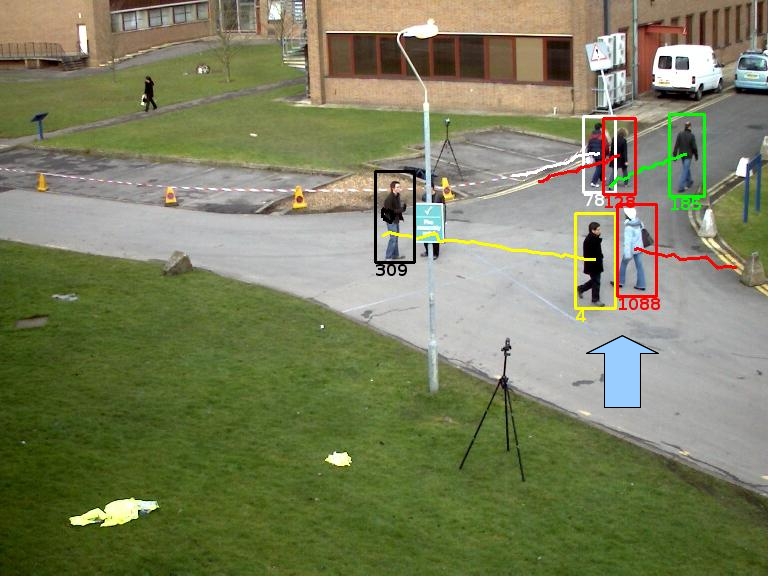} &
\includegraphics[trim=0 100 0 55, clip, width=5cm]{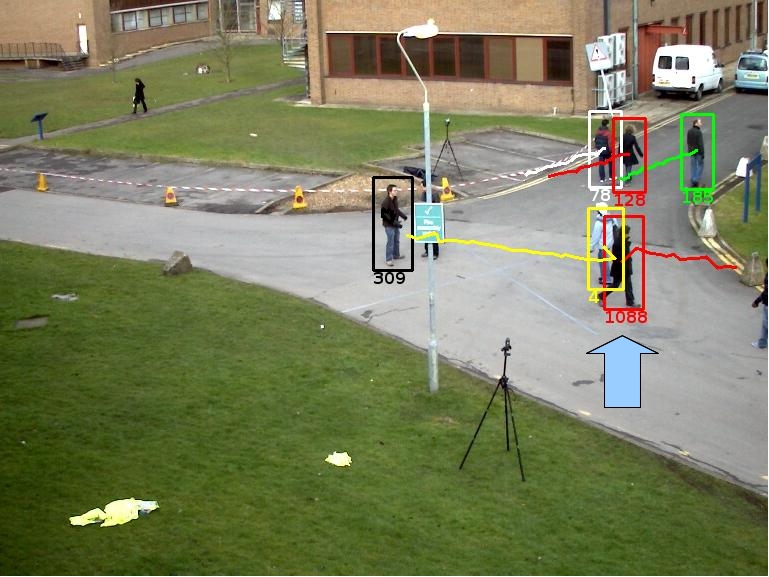} &
\includegraphics[trim=0 100 0 55, clip, width=5cm]{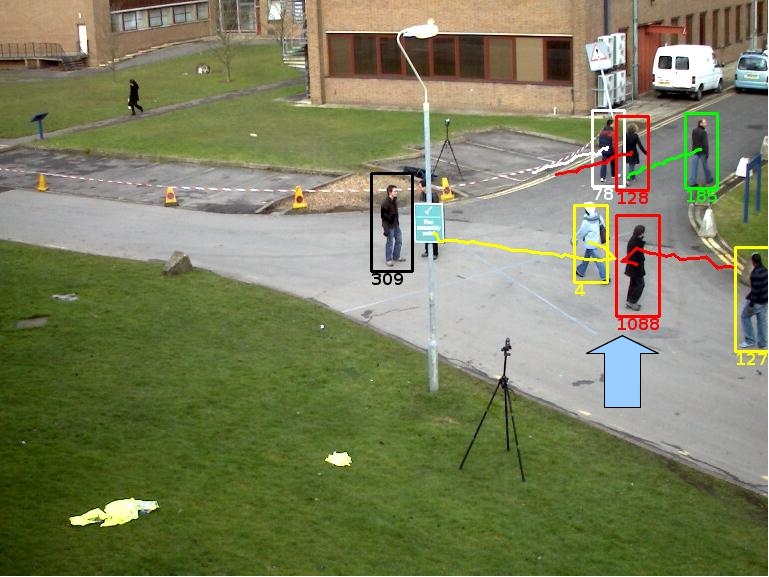} \\
\end{array}$
\end{center}
\caption{\label{fig_pets1_klt} KLT tracker: Two persons of Id 4 and Id 1088 (marked by the cyan arrow) switch their ids after occlusion.}
\end{figure*}

\begin{figure*}[]
\begin{center}$
\begin{array}{ccc}
\includegraphics[width=5cm]{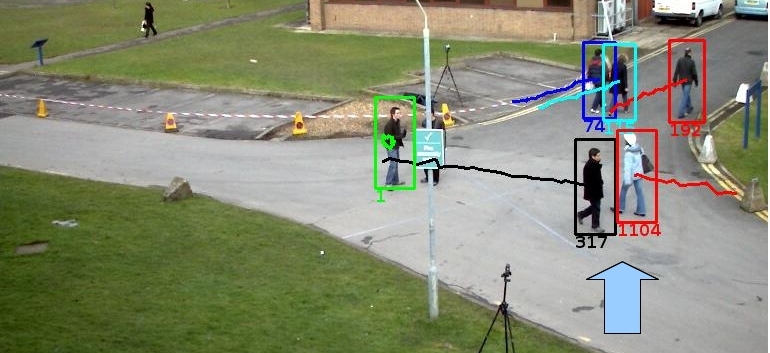} &
\includegraphics[width=5cm]{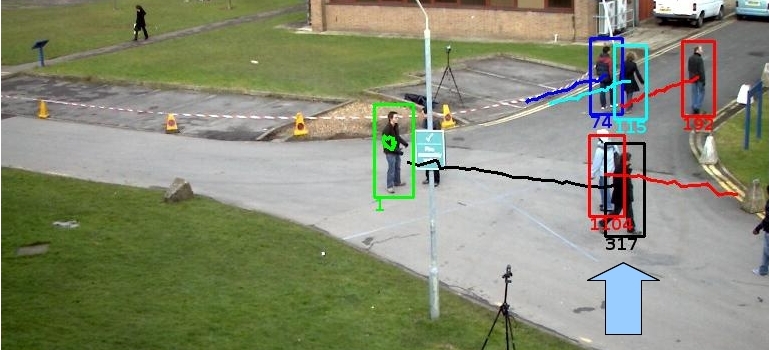} &
\includegraphics[width=5cm]{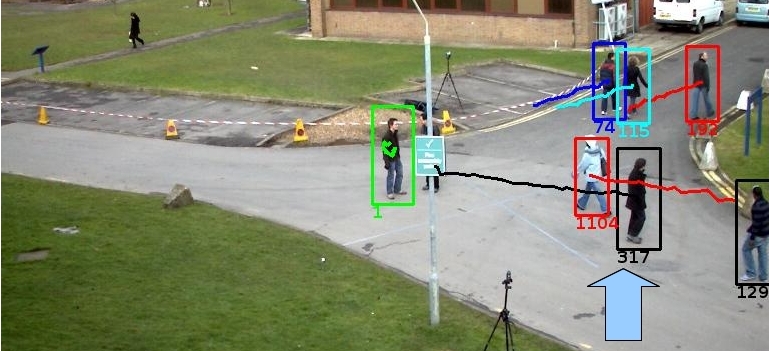} \\
\end{array}$
\end{center}
\caption{\label{fig_pets1_app} Discriminative Appearance-based tracker: Two persons of Ids 317 and 1104 (marked by the cyan arrow) keep correctly their ids after occlusion as they have very different color appearances.}
\end{figure*}

Figures \ref{fig_pets1_klt} and \ref{fig_pets1_app} illustrates the tracking results of the KLT tracker and the appearance-based tracker when an object occlusion occurs (persons id 4 and id 1088 in the figure \ref{fig_pets1_klt}). The proposed approach selects the appearance-based tracker for computing the trajectory of these two objects. While the KLT tracker cannot keep correctly the object Ids after the occlusion, the appearance tracker can track correctly these two objects as they have very different color appearances.  

\begin{figure*}[]
\begin{center}$
\begin{array}{ccc}
\includegraphics[trim=0 120 0 55, clip, width=5cm]{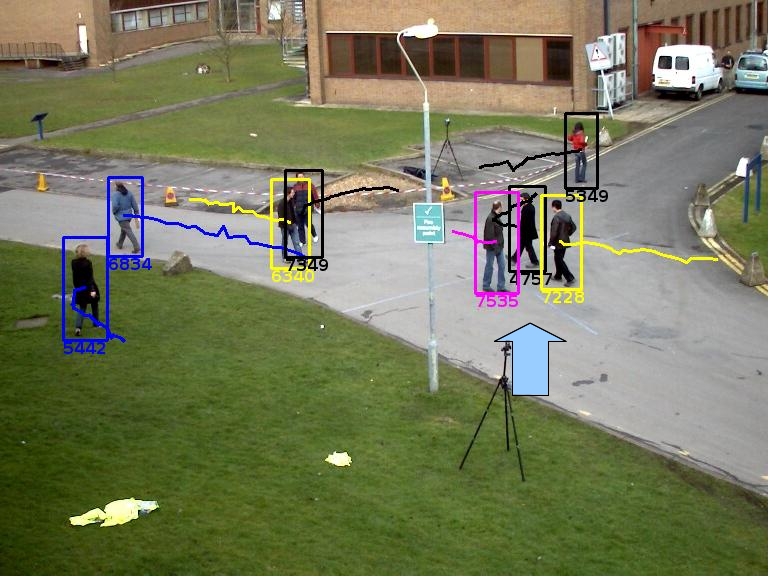} &
\includegraphics[trim=0 120 0 55, clip, width=5cm]{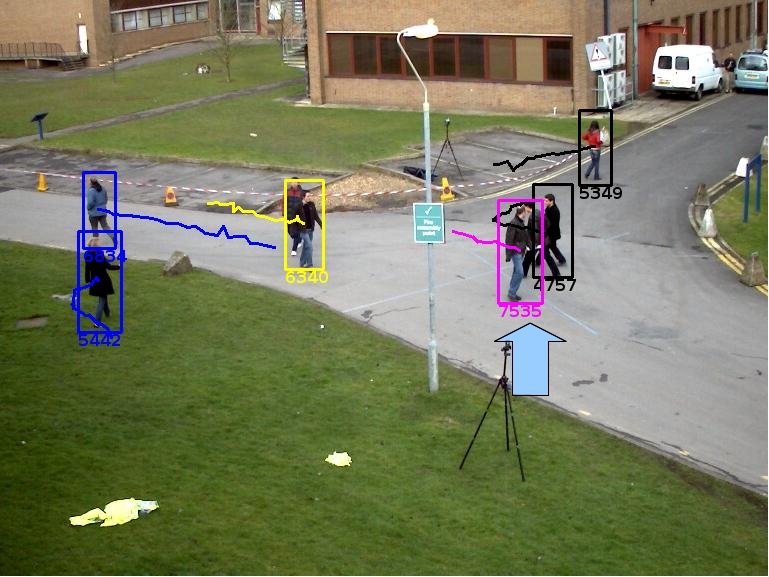} &
\includegraphics[trim=0 120 0 55, clip, width=5cm]{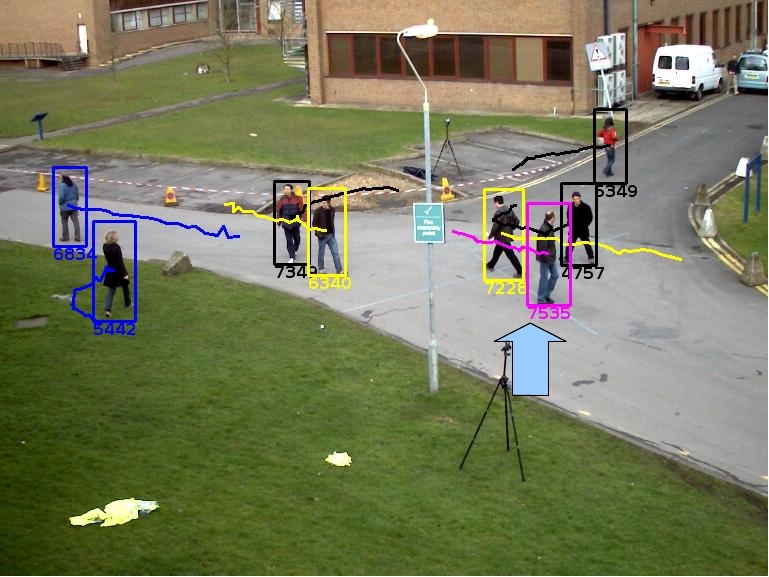} \\
\end{array}$
\end{center}
\caption{\label{fig_pets2_klt} KLT Tracker result: Three persons of Ids 7535, 7228 and 4757 (marked by the cyan arrow) are occluded each other but their identities are kept correctly after occlusion.}
\end{figure*}

\begin{figure*}[]
\begin{center}$
\begin{array}{ccc}
\includegraphics[trim=0 120 0 55, clip, width=5cm]{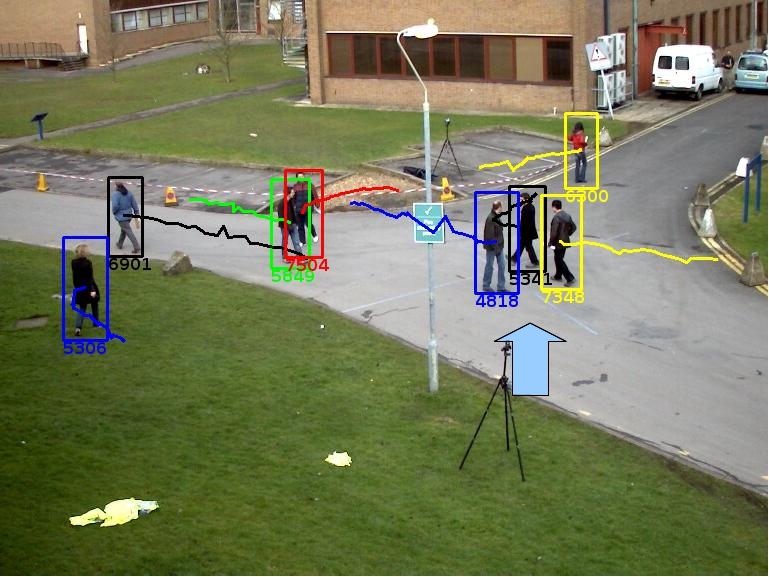} &
\includegraphics[trim=0 120 0 55, clip, width=5cm]{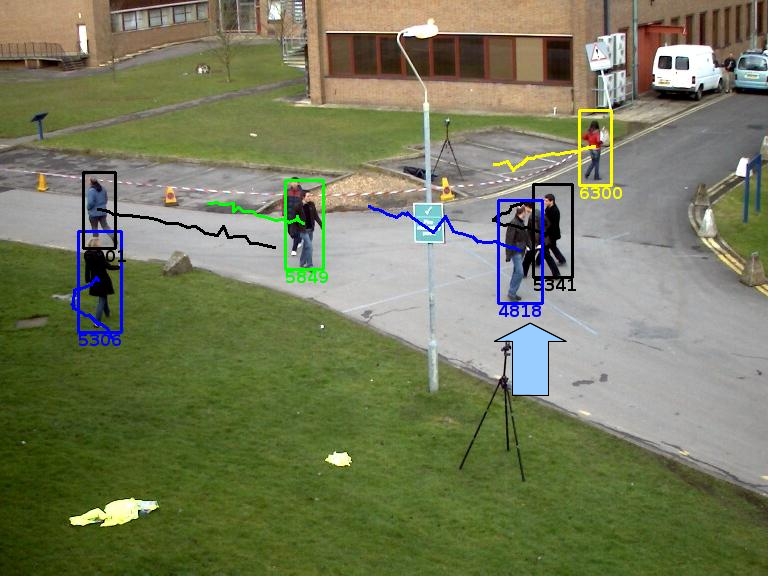} &
\includegraphics[trim=0 120 0 55, clip, width=5cm]{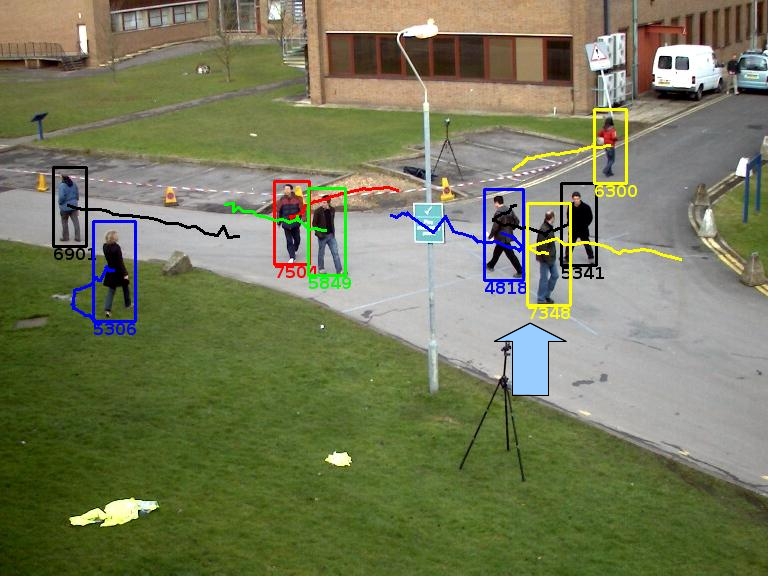} \\
\end{array}$
\end{center}
\caption{\label{fig_pets2_app} Appearance-based Tracker result: Three persons of Ids 4818, 5341 and 7348 (marked by the cyan arrow) are occluded each other. The persons 4818 and 7348 switch their identities after occlusion due to their occlusion and quite similar appearances.}
\end{figure*}

Inversely, in the figures \ref{fig_pets2_klt} and \ref{fig_pets2_app}, the proposed approach selects the KLT tracker for handling the occlusion of persons 7535, 7228 and 4757 (see figure \ref{fig_pets2_klt}). In this case, the objects have quite similar appearances and are occluded hardly. Therefore the appearance tracker fails but the KLT tracker can still keep correctly the person identities.

Table \ref{tab_pets_result} presents the metric results of the proposed approach, the KLT tracker, the appearance tracker and different trackers from the state of the art.  The metric $\overline{M}$ represents the average value of MOTA and MOTP. The result of \cite{shitrit11} is provided by \cite{zamir10}. While using separately the KLT tracker or the appearance tracker, the performance is lower than other approaches in state of the art. The proposed approach by combining these two trackers improves significantly the tracking performance and obtains the best values for all three metrics.

\begin{table}[]
   \begin{center}
	\begin{tabular}{|l|c|c|c|c|}
		\hline
			  Method 	& MOTA & MOTP & $\overline{M}$   \\
 		\hline
	
	 	Berclaz et al. \cite{berclaz06}  	& 0.80	& 0.58 	& 0.69	\\
		\hline
		 Shitrit et al. \cite{shitrit11}  	& 0.81	& 0.58 	& 0.70	\\
		\hline
		 KLT tracker & 0.41	& 0.76 	& 0.59	\\
		\hline
		 Appearance tracker & 0.62	& 0.63 	& 0.63	\\
		\hline
\textbf{Proposed approach}	& \textbf{\textcolor{red}{0.86}} & \textbf{\textcolor{red}{0.72}} & \textbf{\textcolor{red}{0.79}}  \\
		\hline
	\end{tabular}
\end{center}
\caption{\label{tab_pets_result}Tracking results on the PETS sequence S2.L1, camera view 1, sequence time 12.34. The best values are printed in \textbf{\textcolor{red}{red}}.}
\end{table}

\subsection{Caviar Dataset}

In this dataset, we select the tracking evaluation metrics proposed in \cite{li09}. Let $GT$ be the number of trajectories in the ground-truth of the test video. The first metric \textbf{$MT$} computes the number of trajectories successfully tracked for more than 80\% divided by GT. The second metric \textbf{$PT$} computes the number of trajectories that are tracked between 20\% and 80\% divided by GT. The last metric \textbf{$ML$} is the percentage of the left trajectories.

The processing Caviar dataset has 26 sequences. For fair comparison with other approaches, 20 sequences including 143 mobile objects are selected for testing. Table \ref{tab_caviar_result} presents the tracking results of the proposed approach, the KLT tracker, the appearance-based tracker and of some recent trackers from the state of the art. Compared to the performance of KLT and appearance trackers, the proposed approach increases significantly the $MT$ value and decreases the $ML$ value. Our approach gets the best $MT$ value compared to the other trackers.

\begin{table}[]
   \begin{center}
	\begin{tabular}{|l|c|c|c|c|}
		\hline
			  Method 	& MT (\%)& PT (\%) & ML (\%)   \\
 		\hline
  Xing et al. \cite{xing09}		& 84.3 	& 12.1 & 3.6	\\
		\hline
      Li et al. \cite{li09}		& 84.6 	& 14.0 & 1.4  	\\
		\hline
Kuo et al.	 \cite{kuo}		& 84.6 	& 14.7 & \textcolor{red} {\textbf{0.7}}	\\
		\hline
KLT Tracker & 74.4    & 13.4 	& 12.2	\\
		\hline
Appearance Tracker  			& 78.3 	& 16.0 & 5.7	\\
		\hline
\textbf{Proposed approach}		& \textbf{\textcolor{red}{86.4}} & \textbf{10.6} & \textbf{3.0} \\
		\hline

	\end{tabular}
\end{center}
\caption{\label{tab_caviar_result}Tracking results on the Caviar dataset. The best values are printed in \textbf{\textcolor{red}{red}}.}
\end{table}

\subsection{TUD Dataset}

\begin{figure}[t]
\begin{center}
   \includegraphics[trim = 0 130 0 35, clip, width=0.8\linewidth]{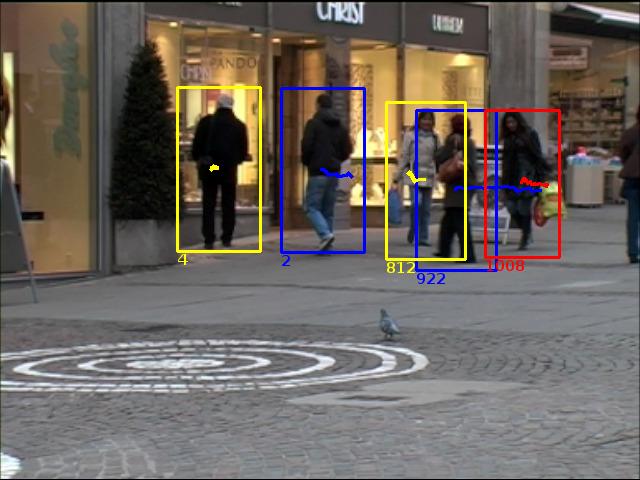}
\end{center}
   \caption{TUD video}
\label{fig_tud}
\end{figure}

For the TUD dataset, we select the TUD-Stadtmitte sequence. This video contains only 179 frames and 10 objects but is very challenging due to heavy and frequent object occlusions (see figure \ref{fig_tud}). Table \ref{tab_tud_result} presents the tracking results of different trackers. Result of \cite{kuo11} is provided by \cite{yang12}. Compared to the KLT and appearance trackers, the proposed approach increases the $MT$ value and decreases the $ML$ value. Our approach obtains the best $MT$ and $ML$ values compared to the other trackers.

\begin{table}[]
   \begin{center}
	\begin{tabular}{|l|c|c|c|c|}
		\hline
			  Method 	& MT (\%)& PT (\%) & ML (\%)   \\
 		\hline
Andriyenko et al. \cite{andriyenko11}		& 60.0 	& 30.0 		& \textbf{\textcolor{red}{10}}  	\\
		\hline
  Kuo et al.	 \cite{kuo11}		& 60.0 	& 30.0 		& \textbf{\textcolor{red}{10}}  	\\
		\hline
KLT Tracker 				& 60.0    & 20.0 	& 20.0	\\
		\hline
Appearance Tracker  			& 50.0 	& 30.0 & 20.0	\\
		\hline
\textbf{Proposed approach}		& \textbf{\textcolor{red}{70.0}} & \textbf{20.0} & \textbf{ \textcolor{red}{10.0}} \\
		\hline
	\end{tabular}
\end{center}
\caption{\label{tab_tud_result}Tracking results on the TUD-Stadtmitte sequence. The best values are printed in \textbf{\textcolor{red}{red}}.}
\end{table}

\section{Conclusion}

We presented in this paper a new object tracking process which is able to cope with video content variation. Object detection is evaluated online and corrected using a KLT feature tracker. The object trackers which are based on KLT and discriminative appearance, are then selected for each object to ensure correct object links over time. The proposed approach has been experimented on three public datasets (PETS, Caviar and TUD). The experimental results show that the proposed method gets much better performance than the KLT and appearance trackers. This approach also outperforms recent state of the art trackers. In the future work, we add more trackers in the selection process to obtain a more robust tracking quality.

\section*{Acknowledgments}
\noindent This work is supported by The Panorama and Support projects.

\balance

{\small
\bibliographystyle{ieee}
\bibliography{egbib}

\begin{thebibliography}{10}\itemsep=-1pt

\bibitem{andriyenko11}
A.~Andriyenko and K.~Schindler.
\newblock Multi-target tracking by continuous energy minimization, 2011.
\newblock In CVPR.

\bibitem{berclaz06}
J.~Berclaz, F.~Fleuret, E.~Turetken, and P.~Fua.
\newblock Multiple object tracking using k-shortest paths optimization.
\newblock {\em TPAMI}, 33(9):1806--1819, 2011.

\bibitem{corvee10}
E.~Corvee and F.~Bremond.
\newblock Body parts detection for people tracking using trees of histogram of
  oriented gradient descriptors.
\newblock In {\em AVSS}, August 2010.

\bibitem{forstner99}
W.~Forstner and B.~Moone.
\newblock A metric for covariance matrices, 1999.
\newblock In Quo vadis geodesia...?, Festschrift for Erik W.Grafarend on the
  occasion of his 60th birthday, TR Dept. of Geodesy and Geoinformatics,
  Stuttgart University (Germany).

\bibitem{kasturi09}
R.~Kasturi, P.~Soundararajan, J.~Garofolo, R.~Bowers, and V.~Korzhova.
\newblock Framework for performance evaluation of face, text, and vehicle
  detection and tracking in video: Data, metrics, and protocol.
\newblock {\em TPAMI}, 31(2):319 -- 336, 2009.

\bibitem{kuo}
C.~Kuo, C.~Huang, and R.~Nevatia.
\newblock Multi-target tracking by online learned discriminative appearance
  models, 2010.
\newblock In CVPR.

\bibitem{kuo11}
C.-H. Kuo and R.~Nevatia.
\newblock How does person identity recognition help multi-person tracking?,
  2011.
\newblock In CVPR.

\bibitem{li09}
Y.~Li, C.~Huang, and R.~Nevatia.
\newblock Learning to associate: Hybridboosted multi-target tracker for crowded
  scene, 2009.
\newblock CVPR.

\bibitem{yangCol}
N.C.Yang, W.H.Chang, C.M.Kuo, and T.H.Li.
\newblock {A fast mpeg-7 dominant color extraction with new similarity measure
  for image retrieval}.
\newblock {\em {J. Visual Communication and Image Representation}}, 2008.

\bibitem{prost}
J.~Santner, C.~Leistner, A.~Saffari, T.~Pock, and H.~Bischof.
\newblock Prost: Parallel robust online simple tracking, 2010.
\newblock In CVPR.

\bibitem{sbak}
S.Bak, E.Corvee, F.Bremond, and M.Thonnat.
\newblock {Person Re-identification Using Spatial Covariance Regions of Human
  Body Parts}.
\newblock In {\em {AVSS}}, 2010.

\bibitem{shi94}
J.~Shi and C.~Tomasi.
\newblock Good features to track, 1994.
\newblock In CVPR.

\bibitem{shitrit11}
H.~B. Shitrit, J.~Berclaz, F.~Fleuret, and P.~Fua.
\newblock Tracking multiple people under global appearance constraints, 2011.
\newblock In ICCV.

\bibitem{xing09}
J.~Xing, H.~Ai, and S.~Lao.
\newblock Multi-object tracking through occlusions by local tracklets filtering
  and global tracklets association with detection responses, 2009.
\newblock CVPR.

\bibitem{yang12}
B.~Yang and R.~Nevatia.
\newblock An online learned crf model for multi-target tracking, 2012.
\newblock In CVPR.

\bibitem{yoon12}
J.~H. Yoon, D.~Y. Kim, and K.-J. Yoon.
\newblock Visual tracking via adaptive tracker selection with multiple
  features, 2012.
\newblock In ECCV.

\bibitem{zamir10}
A.~Zamir, A.~Dehghan, and M.~Shah.
\newblock Gmcp-tracker: Global multi-object tracking using generalized minimum
  clique graphs.
\newblock In {\em ECCV}, 2012.

\end{thebibliography}
}

\end{document}